%% file: example_paper.tex
\documentclass{article}

\usepackage{microtype}
\usepackage{graphicx}
\usepackage{subfigure}
\usepackage{booktabs} 

\usepackage{hyperref}

\usepackage[accepted]{icml2024}

\usepackage{amsmath}
\usepackage{amssymb}
\usepackage{mathtools}
\usepackage{amsthm}

\usepackage[capitalize,noabbrev]{cleveref}

\usepackage{enumitem}

\theoremstyle{plain}

\theoremstyle{definition}

\theoremstyle{remark}

\usepackage[textsize=tiny]{todonotes}

\icmltitlerunning{Clean-image Backdoor Attacks}

\begin{document}

\twocolumn[
\icmltitle{Clean-image Backdoor Attacks}



\renewcommand{\thefootnote}{\textasteriskcentered}
\begin{icmlauthorlist}
\icmlauthor{Dazhong Rong}{zju,ant}
\icmlauthor{Guoyao Yu}{zju}
\icmlauthor{Shuheng Shen}{ant}
\icmlauthor{Xinyi Fu}{ant}
\icmlauthor{Peng Qian}{zju}\\
\footnotemark\icmlauthor{Jianhai Chen}{zju}
\icmlauthor{Qinming He}{zju}
\icmlauthor{Xing Fu}{ant}
\icmlauthor{Weiqiang Wang}{ant}
\end{icmlauthorlist}

\icmlaffiliation{zju}{College of Computer Science and Technology, Zhejiang University, Hangzhou, China}
\icmlaffiliation{ant}{Ant Group, Hangzhou, China}

\icmlcorrespondingauthor{Jianhai Chen}{chenjh919@zju.edu.cn}

\icmlkeywords{Computer Vision, Backdoor Attack}

\vskip 0.3in
]


\footnotetext{College of Computer Science and Technology, Zhejiang University, Hangzhou, China. $^2$Ant Group, Hangzhou, China. Correspondence to: Jianhai Chen $<$chenjh919@zju.edu.cn$>$.}


\begin{abstract}
To gather a significant quantity of annotated training data for high-performance image classification models, numerous companies opt to enlist third-party providers to label their unlabeled data. 
This practice is widely regarded as secure, even in cases where some annotated errors occur, as the impact of these minor inaccuracies on the final performance of the models is negligible and existing backdoor attacks require attacker's ability to poison the training images. 
Nevertheless, in this paper, we propose clean-image backdoor attacks which uncover that backdoors can still be injected via a fraction of incorrect labels without modifying the training images.
Specifically, in our attacks, the attacker first seeks a trigger feature to divide the training images into two parts: those with the feature and those without it.
Subsequently, the attacker falsifies the labels of the former part to a backdoor class.
The backdoor will be finally implanted into the target model after it is trained on the poisoned data.
During the inference phase, the attacker can activate the backdoor in two ways: slightly modifying the input image to obtain the trigger feature, or taking an image that naturally has the trigger feature as input.
We conduct extensive experiments to demonstrate the effectiveness and practicality of our attacks.
According to the experimental results, we conclude that our attacks seriously jeopardize the fairness and robustness of image classification models, and it is necessary to be vigilant about the incorrect labels in outsourced labeling.
\end{abstract}

\input{section/introduction}

\input{section/related-work}
\input{section/preliminaries}
\input{section/our-attacks}
\input{section/experiments}
\input{section/conclusion}


\bibliography{example_paper}
\bibliographystyle{icml2024}

\end{document}

%% file: section/introduction.tex
\section{Introduction}
The field of deep learning has witnessed a rapid growth, which has led many companies to gather their own business data to train deep learning models.
However, these data are often unlabeled, and manually labeling them can be a cumbersome and time-consuming task.
As a result, companies may opt to outsource the data labeling process.
In such scenarios, although the outsourced labels are untrusted and some of them may be incorrect, the training data is trusted and clean.
Therefore, it is commonly considered safe from backdoor attacks.
To explore potential security threats posed by outsourced labels, in this paper we propose clean-image backdoor attacks, focusing on general image classification tasks.
In our attacks, the attacker cannot modify the training images and can only tamper with a limited proportion of the training labels.
Through the poisoned labels, the attacker aims to implant the backdoor into the target model.

From the perspective of shortcut learning~\cite{geirhos2020shortcut,luo2021rectifying}, the backdoor attacks are the attacks which induce the target model to take the shortcut designed by the attacker.
Up to now various sorts of backdoor attacks have been studied, including visible attacks, invisible attacks and clean-label attacks.
However, all present forms of backdoor attacks involve embedding a trigger pattern into the training images to establish the shortcut, rendering them ineffective in our clean-image context.

To execute clean-image backdoor attacks, we shift the thinking of existing methods.
Instead of purposely poisoning training images to create a shortcut, we try to seek out a shortcut which is naturally present within clean training images.
Specifically, our approach involves designing a trigger feature that can categorize training images into two groups: those with and those without the trigger feature.
We then alter the labels of training images with the trigger feature to a backdoor class and feed the corrupted data to the target model.
During the inference phase, the backdoor within the target model can be activated by inputting the images featuring the designated trigger, which results in the output of the backdoor class.
There exist two cases here.
The first is the input images inherently have the trigger feature, and the second is the attacker imperceptibly modifies the input images to bestow the trigger feature on them.

In designing the trigger feature, there are three primary challenges as follows.
The first challenge is ensuring that the proportion of training images with the trigger feature aligns exactly with the poisoning ratio.
Secondly, the feature pattern must have priority in the learning process of the target model.
Failure to do so may result in the target model taking other unexpected shortcuts, rendering the attacker unable to manipulate the backdoor with precision.
Finally, to enable activation of the backdoor via the poisoned images which are slightly modified by the attacker from the clean images during the target model's inference stage, the feature must be sensitive to minor and imperceptible modifications.
This sensitivity ensures that any clean image without the trigger feature can be transformed into an image featuring the designated trigger with minimal effort.

To address the above challenges, our designed trigger feature is based on a kernel-based linear scoring function, which can ensure the satisfaction of the first two points.
For the third point, we propose two attack strategies resulting in different levels of the feature sensitivity.
Specifically, in the first strategy of randomizing, the kernel matrix of our linear scoring function is randomly initialized with the standard gaussian distribution.
In the second strategy of learning, the kernel matrix is further optimized by gradient descent for better feature sensitivity.

The key contributions of our work can be summarized as the following three points:
\begin{itemize}[leftmargin=*]
	\item To the best of our knowledge, our proposal of clean-image backdoor attacks is the first of its kind to apply invisible backdoor attacks on the general image classification task, without requiring the modification of training images during the training stage.
	Specifically, we provide a formal definition of clean-image backdoor attacks and propose two metrics (\textit{i.e.,} natural and manual attack success rates) for evaluating attack effectiveness.
	\item We define a metric that indicates feature sensitivity in our attacks and propose two strategies (randomizing and learning) for constructing the trigger feature, which yields different levels of feature sensitivity.
	\item We conduct extensive experiments on two popular image classification datasets using a variety of target models, including fully-connected based, CNN based, and Transformer based models.
	In our experimental section, we demonstrate the practicality of our attacks from two perspectives (effectiveness and stealthiness) and provide a deep analysis of the difference in attack performance resulting from the use of the two aforementioned strategies.
\end{itemize}

%% file: section/related-work.tex
\section{Related Work}
Backdoor attacks have garnered considerable attention as a rapidly growing research area due to their ability to expose vulnerabilities in deep neural networks (DNNs).
In this section, we present a comprehensive overview of the development of backdoor attacks in the context of general image classification tasks and categorize existing studies into four distinct branches following~\cite{li2023backdoor,guo2022overview,li2022backdoor}.
In the subsequent subsections, we provide a detailed explanation of each of these four types of backdoor attacks in the order of their appearance. 
From our discussion, we can summarize the research trend in the field of backdoor attacks that the capability restrictions on attackers are getting more and more stringent.

\subsection{Visible Backdoor Attacks}
In~\cite{gu2019badnets}, the researchers propose a representative visible backdoor attack named BadNet.
This attack assumes that the training of the target model is outsourced to the attacker, who can manipulate the training freely.
Specifically, the attacker randomly selects a subset of training samples and creates backdoored versions of these samples by adding a trigger to the original clean images, such as a pattern of bright pixels in the bottom right corner.
The labels of the backdoored samples are set by the attacker to achieve their goal.
The attacker then includes these poisoned samples in the training set used to train the target model.
As a result, the target model becomes backdoored.
When the target model encounters an image with the backdoor trigger, it predicts the attacker's desired label as the output.
In all other cases, the model works properly as expected.

Several studies~\cite{shokri2020bypassing,lin2020composite,nguyen2020input,xue2020one,gong2021defense} have built upon the work of BadNet and attempted to improve it. 
These studies have focused on making the backdoors in the target models more difficult to detect, or enabling the backdoors to be activated more flexibly. 
However, it is important to note that all of these studies do not restrict the attacker's ability to modify the training data.

\subsection{Invisible Backdoor Attacks}
In situations where the attacker lacks the ability to manipulate the training of the target model, and is only able to provide training data, existing visible backdoor attacks become impractical as their poisoned images can be easily identified by the human eye. 
To overcome this limitation, the first invisible backdoor attack was introduced in~\cite{chen2017targeted}, which involves generating a poisoned image that is a blend of the original clean image and the backdoor trigger. 
In order to maintain imperceptibility, the blend ratio is carefully controlled to ensure that the poisoned image is visually indistinguishable from the clean image.

Subsequently, numerous other invisible backdoor attacks have been proposed in the literature~\cite{li2020invisible,doan2021lira,doan2021backdoor,zhao2022defeat,xia2022enhancing,li2021invisible,zhang2022poison,wang2022bppattack}. 
These works employ a variety of techniques such as steganography and image quantization to create poisoned images that are imperceptible to the human eye. 
In all of these attacks, the attacker is limited to imperceptibly poisoning the training images, but has free reign to manipulate their labels as desired.

\subsection{Clean-label Backdoor Attacks}
The incorrect labels in the above invisible backdoor attacks may be noticed, possibly leading to the detection of the backdoor attacks. 
To further enhance the stealthiness of backdoor attacks, the clean-label backdoor attack was firstly introduced in~\cite{turner2019label}. 
In this approach, the attacker poisons only the training images in the backdoor class using either adversarial perturbations or a generative model. 
As a result, the poisoned images not only appear indistinguishable from their clean counterparts to the human eye, but their labels are also correct with no anomalies.

A considerable number of studies have opted for the setting where the attacker is only able to poison the training images but not their labels. 
In line with this setting, various clean-label backdoor attacks have been proposed~\cite{saha2020hidden,ning2021invisible,gao2023not,liu2020reflection,barni2019new,luo2022enhancing,zeng2022narcissus}. 
However, compared to the backdoor attacks that allow the attacker to poison the labels, the clean-label backdoor attacks are generally less effective, which is reasonable given the harsh restrictions on the attacker's ability. 
Within the clean-label backdoor attacks, a higher poisoning rate is required but a lower attack success rate is achieved.

\subsection{Clean-image Backdoor Attacks}
Although the restrictions in clean-label backdoor attacks are already extremely harsh, these attacks are still infeasible in some scenarios.
This is especially true when the training data for the target model is collected by its holder, rendering attackers incapable of poisoning the training images to their advantage.
For instance, some companies gather a vast number of unlabeled images from their business and outsource the labeling of these images to train their target models.
In such cases, all existing backdoor attacks are rendered completely ineffective.
In light of this, we propose to modify the restrictions such that the attacker can only poison a small fraction of labels in the training data while leaving the training images untouched. 
We refer to these attacks as clean-image backdoor attacks.

To the best of our knowledge, it is widely believed that injecting a backdoor without poisoning the training images is impossible.
There exists only one study~\cite{chenclean} that attempts to execute clean-image attacks on DNNs.
This work restricts its scope to multi-label image classification tasks, in which the attacker selects a specific combination of categories (such as \{pedestrian, car, traffic light\}) as the trigger pattern and falsifies the labels of training images containing the category combination to implant the backdoor. 
However, we argue that such attacks are not traditional backdoor attacks because the attacker is unable to activate the backdoor by generating poisoned images based on clean images after poisoning the training.
In other words, the backdoor activation in these attacks relies on clean images naturally containing the trigger pattern, which means the attacker cannot manipulate it freely.
Additionally, these attacks are not generic due to the category combination triggers are completely unsuitable for single-label scenarios.

%% file: section/preliminaries.tex
\section{Preliminaries}
In this section, we first formulate the problem of our attacks and then introduce relevant limitations.

\subsection{Problem Formulation}~\label{sec:3.1}
Our clean-image attacks aim to backdoor the general image classification models.
Specifically, we denote the original clean training set as $\mathcal{D}^{train}$.
Each sample $(\mathbf{X}, y)\in\mathcal{D}^{train}$ comprises an image $\mathbf{X}\in\mathcal{X}=\mathbb{R}^{C\times W\times H}$ and its label $y\in\mathcal{Y}=\{1,2\dots,N\}$, where $C$ represents the number of channels, $W$ and $H$ represent the width and height of the images respectively, and $N$ is the number of classes.
The backdoor class designated by the attacker is denoted as $y^T\in\mathcal{Y}$.
There are two phases in our attacks as follows.

\textbf{Label poisoning phase.} 
Prior to training the target image classification model, the attacker initiates the poisoning phase. 
A specific subset of samples represented by $\mathcal{D}_p$ is intentionally selected from the pristine training set $\mathcal{D}^{train}$, and their labels are falsified to the backdoor class $y^T$.
Consequently, the training set becomes poisoned and can be denoted as $\mathcal{D}^{train}_p=(\mathcal{D}^{train}\setminus\mathcal{D}_p)\bigcup\{(\mathbf{X}, y^T) | (\mathbf{X}, y)\in\mathcal{D}_p\}$.
Following this, the target model is trained on the poisoned training set $\mathcal{D}^{train}_p$, and thereby the backdoor is implanted.

\textbf{Trigger adding phase.}
During the inference stage, for any clean image $\mathbf{X}$ whose ground-truth label is not equal to $y^T$, the attacker can slightly modify it to a poisoned image $\mathbf{t}(\mathbf{X})$ (or simply taking $\mathbf{t}(\mathbf{X})=\mathbf{X}$ if the clean image can naturally activate the backdoor).
The well-crafted poisoned image is highly similar to the original clean image, but with the backdoor trigger embedded within it.
Upon passing the poisoned image as input to the target model, the hidden backdoor within the model is activated, resulting in the prediction of the backdoor class $y^T$ as the output.

It is noteworthy that our attack methodology has the following characteristics:
(i) The attacker is restricted to falsifying the labels in the training set, and cannot manipulate the training of the target model in any other manner, such as adding extra training samples or modifying the batch order.
(ii) Our attacks are black-box attacks, as the attacker lacks any prior knowledge regarding the target model structure.

\subsection{Attack Limitations}~\label{sec:3.2}
Considering the stealthiness and the practicality of our attacks, there are mainly two limitations on our attacks:

\textbf{Poisoning ratio.}
In the label poisoning phase, we define the poisoning ratio $\beta$ as the proportion of the poisoned samples in the training set (\textit{i.e.,} $\beta=\frac{|\mathcal{D}_p|}{|\mathcal{D}^{train}|}$), and restrict the value of $\beta$.
A larger $\beta$ means a larger number of incorrect labels in the final training set $\mathcal{D}^{train}_p$.
It is suspicious if the data quality of $\mathcal{D}^{train}_p$ is low.
Besides, too many incorrect labels cause severe drop in the performance of the target model.

\textbf{Difference norm.}
In the trigger adding phase, we expect the difference between the clean image and the generated poisoned image is imperceptible, or it may lead to the poisoned image being noticed by the naked eye.
Therefore, in our attacks we restrict the $\ell_2$-norm of the difference to be no greater than a hyper-parameter $\lambda$ (\textit{i.e.,} $\|\mathbf{t}(\mathbf{X})-\mathbf{X}\|_2\le\lambda$).

%% file: section/our-attacks.tex
\section{Clean-image Backdoor Attacks}
In this section, we present our proposed clean-image backdoor attacks in detail.
Specifically, we first introduce the core idea and the challenges of our attacks.
Then we describe the design of the binary classification function in our attacks.
At last, we demonstrate our attack workflows.

\subsection{Core Idea and Challenges}\label{sec:4.1}
Inspired by the recent studies~\cite{geirhos2020shortcut,luo2021rectifying} on shortcut learning, the core idea of our attacks is to induce the target image classification model to take a shortcut.
As a simple example, if in the training set all the images taken on cloudy days are classified to the backdoor class $y^T$, the target model may learn the shortcut pattern: when the input image $\mathbf{X}$ is taken on cloudy days, the output label is $y^T$.
Essentially the learned shortcut is the implanted backdoor in the target model.

More generally, in our attacks we need to find a common feature (\textit{e.g.}, taken on cloudy days), and falsify the labels of all images with this feature in the training set to $y^T$ for creating the shortcut.
We can use a binary classification function $f:\mathcal{X}\rightarrow\{0,1\}$ to describe the common feature.
For the image $\mathbf{X}$ which has the common feature, $f(\mathbf{X})=1$; otherwise, $f(\mathbf{X})=0$.
Let $\mathcal{F}_1=\{\mathbf{X}|(\mathbf{X},y)\in\mathcal{D}^{train}\land f(\mathbf{X})=1\}$ and $\mathcal{F}_0=\{\mathbf{X}|(\mathbf{X},y)\in\mathcal{D}^{train}\land f(\mathbf{X})=0\}$ denote the set of the images in $\mathcal{D}^{train}$ with the common feature and without the common feature, respectively.
The following three points need to be considered when designing the function $f$, which is quite challenging.

\textbf{Ratio Consistence.}
We define the proportion of the images with the common feature as $\gamma=\frac{|\mathcal{F}_1|}{|\mathcal{D}^{train}|}$.
On the one hand, $\gamma$ can not be too large.
In the label poisoning phase of our attacks, the number of the poisoned samples in $\mathcal{D}^{train}_p$ is limited by the poisoning ratio $\beta$.
If $\gamma>\beta$, the attacker can not falsify all the labels of the images with the common feature, which means $\exists(\mathbf{X},y)\in\mathcal{D}^{train}_p,f(\mathbf{X})=1\land y\neq y^T$.
These samples defy the shortcut pattern and hence will weaken the shortcut learning in the target model.
On the other hand, $\gamma$ can not be too small.
A smaller $\gamma$ means a smaller number of the samples in $\mathcal{D}^{train}_p$ which can reflect the shortcut pattern.
It may result in overfitting of the target model for shortcut learning.
In general, ideally $\gamma$ is exactly equal to $\beta$.

\textbf{Shortcut Priority.}
To ensure the target model first learns our desired shortcut, the binary classification function $f$ should not be too complicated, and its pattern can be easily captured by the target image classification model.
Otherwise, the target image classification model may take some other unexpected shortcuts, leading to the fact that the attacker can not precisely control the activation of the backdoor in the trigger adding phase of our attacks.

\textbf{Feature Sensitivity.}
As mentioned in the previous Section~\ref{sec:3.2}, there is a limitation on the difference between the clean images and the poisoned images in the trigger adding phase of our attacks.
Therefore, the common feature we adopt should be sensitive enough.
For any clean image without the common feature, we can turn it to have the common feature with just a little modification.
More specifically, for $\forall(\mathbf{X},y)\in\mathcal{D}^{train}$, the binary classification function $f$ should satisfy: 
\begin{equation}
\max\limits_{||\Delta||_2\le\lambda} \Big\{f(\mathbf{X}+\Delta)\Big\}=1.
\label{eq:1}
\end{equation}

\subsection{Function Design}
To strictly keep the aforementioned ratio consistence, we design a scoring function $g:\mathcal{X}\rightarrow\mathbb{R}$ rather than straightly design the binary classification function $f$.
For $\forall\mathbf{X}\in\mathcal{D}^{train}$, we sort their values of $g(\mathbf{X})$ in descending order.
Then we set the $(\big\lfloor\beta\cdot|\mathcal{D}^{train}|\big\rfloor)$-th value as the threshold $\alpha$.
The function $f$ can be obtained from $g$ by $f(\mathbf{X})=\Gamma\big(g(\mathbf{X})-\alpha\big)$, where $\Gamma$ is the heaviside function: $\Gamma(x)=0$ if $x<0$; $\Gamma(x)=1$ otherwise.
In this way, we can guarantee the ideal value of $\gamma$.

Moreover, taking into account the shortcut priority, we keep the simplicity of the function $g$ from two aspects: 
(i) we limit the number of independent variables in $g$ by making $g$ only related to the bottom-right $S\times S$ squares of the images, where $S$ is a hyper-parameter;
(ii) we restrict $g$ to be completely linear.
Let $\mathbf{X}^i$ denote the $i$-th channel of the image $\mathbf{X}$, where $i\in\{1,2,\dots C\}$ and $\mathbf{X}^i\in\mathbb{R}^{W\times H}$.
Let $\mathbf{X}^{i, S\times S}\in\mathbb{R}^{S\times S}$ denote the bottom-right $S\times S$ square in the matrix $\mathbf{X}^i$.
Finally the definition of our designed function $g$ is as following:
\begin{equation}
	g(\mathbf{X})=\sum_{i=1}^C K^{S\times S}\odot\mathbf{X}^{i,S\times S},
	\label{eq:2}
\end{equation}
where $K^{S\times S}$ is the weight matrix with the shape $S\times S$, and $\odot$ denotes the matrix inner product (\textit{i.e.,} the sum of the element-wise product).
From another perspective, the function $g$ has a similar structure to the convolution function in convolutional neural networks (CNNs)~\cite{albawi2017understanding}.
Further, we can consider the function $f$ based on $g$ as a simplified convolutional layer with $K^{S\times S}$ as the kernel and $-\alpha$ as the bias.
In this simplified convolutional layer, the kernel $K^{S\times S}$ is shared among all channels.
Its activation function is the heaviside function $\Gamma$.
The numbers of its input channels and output channels are $C$ and $1$, respectively.
Note that most existing image classification models are CNN-based, and for these models the pattern of the function $f$ is even easier to capture.

Based on Eq~\ref{eq:1} we can derive that for $\forall (\mathbf{X}, y)\in\mathcal{D}^{train}$: $\max\limits_{\|\Delta\|_2\le\lambda} \Big\{g(\mathbf{X}+\Delta)\Big\}\ge\alpha$.
Since $g$ is a linear function, we can further infer that:
\begin{equation}
	\max\limits_{\|\Delta\|_2\le\lambda} \Big\{g(\Delta)\Big\}\ge\alpha-\min\limits_{(\mathbf{X},y)\in\mathcal{D}^{train}} \Big\{g(\mathbf{X})\Big\}.
	\label{eq:3}
\end{equation}
In the design of the function $g$, we guarantee the feature sensitivity from two sides of the above inequation.
Let $\Delta^{i, S\times S}$ denote the $i$-th channel of the bottom-right $S\times S$ square of $\Delta$. 
For the left side, according to the Cauchy-Schwarz inequality, when $\Delta^{1, S\times S}=\Delta^{2, S\times S}=\dots=\Delta^{C, S\times S}=\frac{\lambda}{\sqrt{C}}\cdot \frac{K^{S\times S}}{\|K^{S\times S}\|_2}$, $g(\Delta)$ takes its maximum value $\frac{\lambda}{\sqrt{C}}\cdot\|K^{S\times S}\|_2$.
And removing the constant, we maximize $\|K^{S\times S}\|_2$.
For the right side, directly minimizing $\alpha$ and maximizing $\min\limits_{(\mathbf{X},y)\in\mathcal{D}^{train}} \big\{g(\mathbf{X})\big\}$ are difficult because both of their values are related to the whole training set.
Given that the right side denotes the gap between the quantile at $\beta$ and the minimum value, it is approximately proportional to the standard deviation $\sigma_g$ of $g(\mathbf{X})$ for all $(\mathbf{X},y)\in\mathcal{D}^{train}$.
Therefore, we turn to minimize $\sigma_g$, which can be easily estimated.
A smaller $\sigma_g$ means less dispersion of $g(\mathbf{X})$ in the training set, hence the high probability of a smaller value of the right side.
Combining the above two perspectives, we define a global metric $\phi_g$ to indicate the feature sensitivity as following: $\phi_g=\frac{\sigma_g}{\|K^{S\times S}\|_2}$.
The smaller $\phi_g$ means the better feature sensitivity.

In our attacks we try two strategies to set the value of the kernel matrix $K^{S\times S}$, which results in different levels of the theoretic feature sensitivity, as following:

\textbf{Randomizing Strategy.} Following~\cite{huang2020instahide}, for $\forall(\mathbf{X},y)\in\mathcal{D}^{train}$, we assume each element in $\mathbf{X}$ obeys a gaussian distribution (\textit{i.e.,} $\forall x\in\mathbf{X}, x\sim\mathcal{N}(\mu,\sigma^2)$). 
Based on this assumption and Eq~\ref{eq:2}, we can infer that $g(\mathbf{X})\sim\sum_{i=1}^C\sum_{k\in K^{S\times S}} k\cdot\mathcal{N}(\mu,\sigma^2)$, and $\sigma_g^2=C\cdot\sum_{k\in K^{S\times S}}k^2\sigma^2$.
Then we have: $\phi_g=\frac{\sqrt{C\cdot\sum_{k\in K^{S\times S}}k^2\sigma^2}}{\|K^{S\times S}\|_2}=\sqrt{C}\cdot\sigma$.
From the above conclusion, we can find that the feature sensitivity is unrelated to the kernel matrix $K^{S\times S}$, which means that the value of $K^{S\times S}$ does not matter.
Hence, in the strategy of randomizing, we directly randomize each element in $K^{S\times S}$ with a random distribution (\textit{e.g.,} the standard gaussian distribution).

\textbf{Learning Strategy.} Due to the fact that the assumption in the randomizing strategy does not exactly match the real training set, there is still room for further improvement.
In the strategy of learning, we first randomly initialize the kernel matrix $K^{S\times S}$, and then repeat a number of rounds to learn its optimal value.
In each round: 
(i) we first randomly select $M$ samples from $\mathcal{D}^{train}$ denoted by $\mathcal{D}'$, where $M$ is a hyper-parameter;
(ii) then we estimate $\sigma_g$ through $\mathcal{D}'$ according to the sample variance formula as:
$\sigma_g=\sqrt{\frac{\sum_{(\mathbf{X},y)\in\mathcal{D}'} (g(\mathbf{X})-\bar{g})^2}{M-1}}\cdot$, where $\bar{g}=\frac{1}{M}\cdot \sum_{(\mathbf{X},y)\in\mathcal{D}'} g(\mathbf{X})$;
(iii) lastly we define the loss as $\phi_g$ with the estimated $\sigma_g$, and use gradient descent to optimize the value of $K^{S\times S}$ for minimizing the loss.
After these rounds of learning, The final kernel matrix $K^{S\times S}$ is obtained.

\begin{figure*}[t]
	\centering
	\includegraphics[width=1\linewidth]{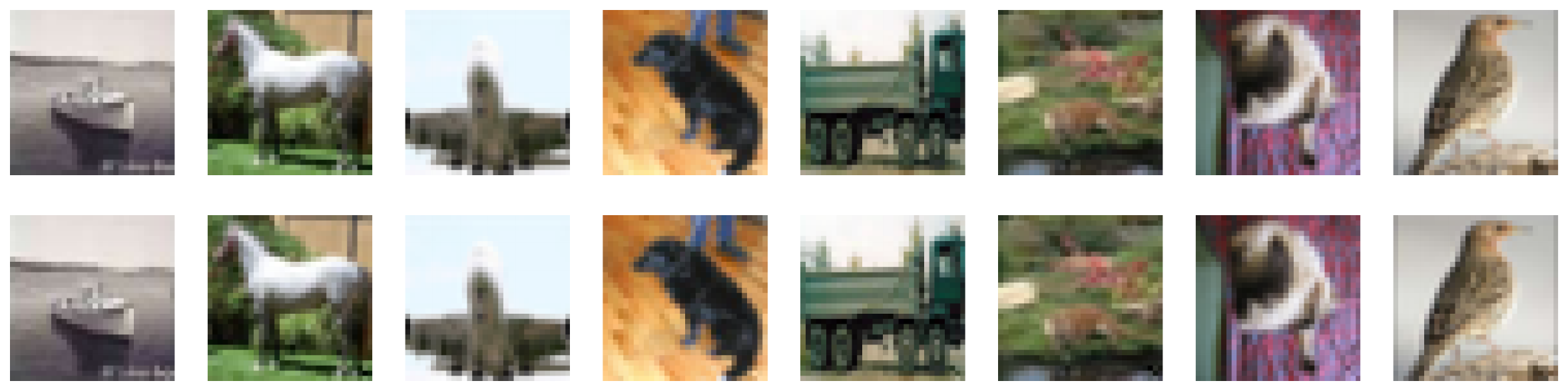}
	\vspace{-20px}
	\caption{Original clean images (top) and generated poisoned images (bottom)}
	\vspace{-10px}
	\label{fig:3}
\end{figure*}

\subsection{Attack Workflows}
With the function $f$ and $g$ are settled, our attacks can be conducted.
Here we introduce the specific workflows of our attacks in two phases respectively.
In the label poisoning phase, the workflow is fairly simple.
What the attacker needs to do is just taking $\mathcal{D}_p=\{(\mathbf{X},y)\in\mathcal{D}^{train}|f(\mathbf{X})=1\}$ and poisoning the labels as we mentioned before in Section~\ref{sec:3.1}.
In the trigger adding phase, for any clean image $\mathbf{X}\in\mathcal{X}$, if $f(\mathbf{X})=1$, it is a natural backdoor image and we do not need to poison it (\textit{i.e.,} $\mathbf{t}(\mathbf{X})=\mathbf{X}$).
If $f(\mathbf{X})=0$, we generate the poisoned image $\mathbf{t}(\mathbf{X})$ as follows.

As aforementioned in Section~\ref{sec:3.2}, we expect the difference between the clean images and the poisoned images is as little as possible.
Hence we can formulate the problem of generating poisoned images as following:
\begin{equation}
	\begin{aligned}
		\min\limits_{\mathbf{t}(\mathbf{X})} \quad&\quad \|\mathbf{t}(\mathbf{X}) - \mathbf{X}\|_2\\
		\textit{s.t.} \quad&\quad g\big(\mathbf{t}(\mathbf{X})\big)\ge\alpha+\delta,
	\end{aligned}
	\label{eq:4}
\end{equation}
where $\delta>0$ is the redundancy to ensure the full activation of the backdoor.
Note that we usually take $\alpha+\delta$ equal to the median of $g(\mathbf{X})$ for $\forall(\mathbf{X},y)\in\mathcal{D}_p$.
Due to the fact that $g$ is only related to the bottom-right $S\times S$ square of the input image, we set the elements in $\mathbf{t}(\mathbf{X})$ equal to the corresponding elements in $\mathbf{X}$ except for the elements in the bottom-right square.

For convenience, we define $\Delta=\mathbf{t}(\mathbf{X})-\mathbf{X}$, and use $\Delta^{i,S\times S}$ to denote the $i$-th channel of the bottom-right square of $\Delta$.
Since our function $g$ is linear, the restriction in Eq~\ref{eq:4} can be transformed to: $g\big(\Delta\big)\ge\alpha+\delta-g(\mathbf{X})$.
Once again according to the Cauchy–Schwarz inequality, we can derive the optimal solution of the above problem as:
$\Delta^{1, S\times S}=\Delta^{2, S\times S}=\dots=\Delta^{C, S\times S}=\frac{\alpha+\delta-g(\mathbf{X})}{C\cdot\|K^{S\times S}\|}\cdot \frac{K^{S\times S}}{\|K^{S\times S}\|_2}$.
Considering the $\ell_2$-norm limitation $\lambda$ and our conclusion on the left side of Eq~\ref{eq:3}, for $\forall i\in\{0,1,\dots,C\}$ we crop $\Delta^{i, S\times S}$ to $V(\frac{\alpha+\delta-g(\mathbf{X})}{C\cdot\|K^{S\times S}\|})\cdot \frac{K^{S\times S}}{\|K^{S\times S}\|_2}$, where $V$ is the cropping function: $V(x)=0$ if $x<0$; $V(x)=\frac{\lambda}{\sqrt{C}}$ if $x>\frac{\lambda}{\sqrt{C}}$; $V(x)=x$ otherwise.

Finally, the generated poisoned image $\mathbf{t}(\mathbf{X})$ can be obtained from the cropped $\Delta$ accordingly.

%% file: section/experiments.tex
\section{Experiments}
In this section, we present our attack performance across the board.
Through our extensive experiments, we mainly aim to investigate the following research questions (RQs):

\textbf{RQ1:} Are our attacks generally effective across various datasets and various target models?

\textbf{RQ2:} What is the impact of the poisoning ratio?

\textbf{RQ3:} How stealthy are our attacks?

\textbf{RQ4:} What is the difference between the two attack strategies (\textit{i.e.,} randomizing and learning)?

\subsection{Experimental Setup}
Here we introduce datasets, target models, evaluation metrics and parameter settings of our experiments, respectively.
 
\textbf{Datasets.} To demonstrate the generality of our attacks, we use two datasets with different complexity: MNIST~\cite{deng2012mnist} and CIFAR10~\cite{krizhevsky2009learning}.
Both of them are publicly accessible and widely used in the field of image classification.
In MNIST, there are $60,000$ samples for training and $10,000$ samples for test.
Each sample consists of a $28\times 28$ grey-scale image of handwritten digits and its label in $0$ to $9$.
In CIFAR10, there are $50,000$ samples for training and $10,000$ samples for test.
Each sample consists of a $32\times 32$ colored image and its class label in $0$ to $9$ respectively denoting airplane, automobile, bird, cat, deer, dog, frog, horse, ship and truck.
Note that in the data preprocessing we normalize the channels of the images.

\textbf{Target Models.} To demonstrate our attacks apply to all sorts of image classification models, we respectively select three representative models as our target models: FCNN for fully-connected based models, ResNet20~\cite{he2016deep} for CNN based models and SimpleViT~\cite{beyer2022better,dosovitskiy2020image} for Transformer based models.
More specifically, our adopted FCNN is a two-layer fully-connected neural network with \textit{hidden\_dim=512}.
Our implementation of ResNet20 strictly follows its original paper.
In SimpleVit, we set: \textit{patch\_size=4}, \textit{dim=32}, \textit{depth=3}, \textit{heads=4} and \textit{mlp\_dim=64}.

\begin{figure*}[t]
	\centering
	\includegraphics[width=0.75\linewidth]{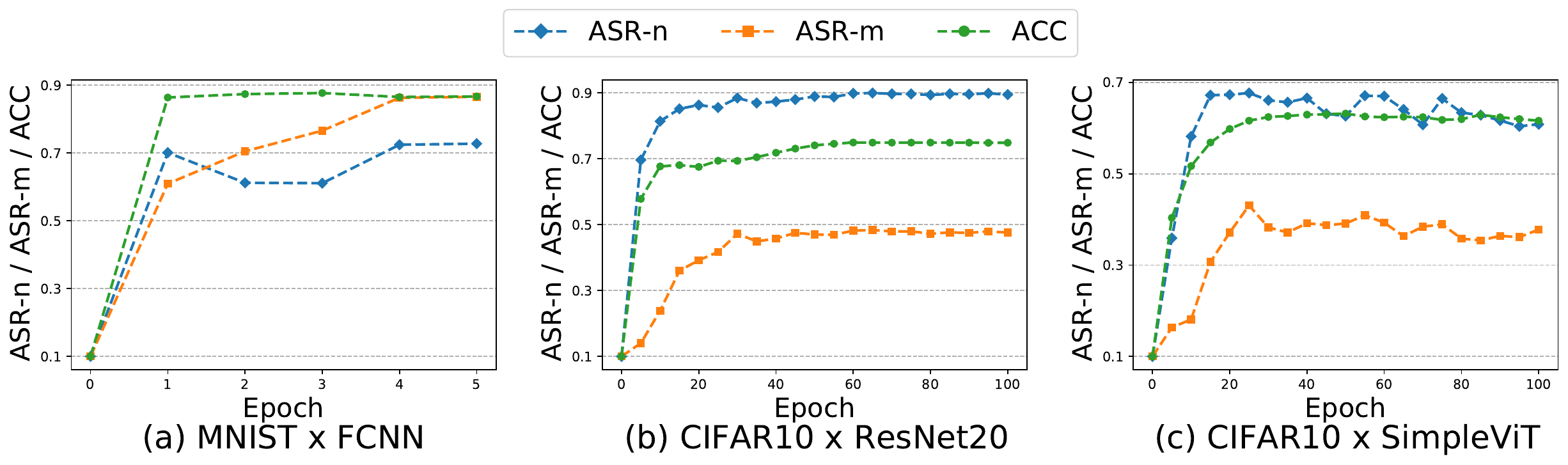} 
	\vspace{-10px}
	\caption{Performance across different datasets and models}
	\label{fig:1}
	\vspace{8px}
	\includegraphics[width=0.75\linewidth]{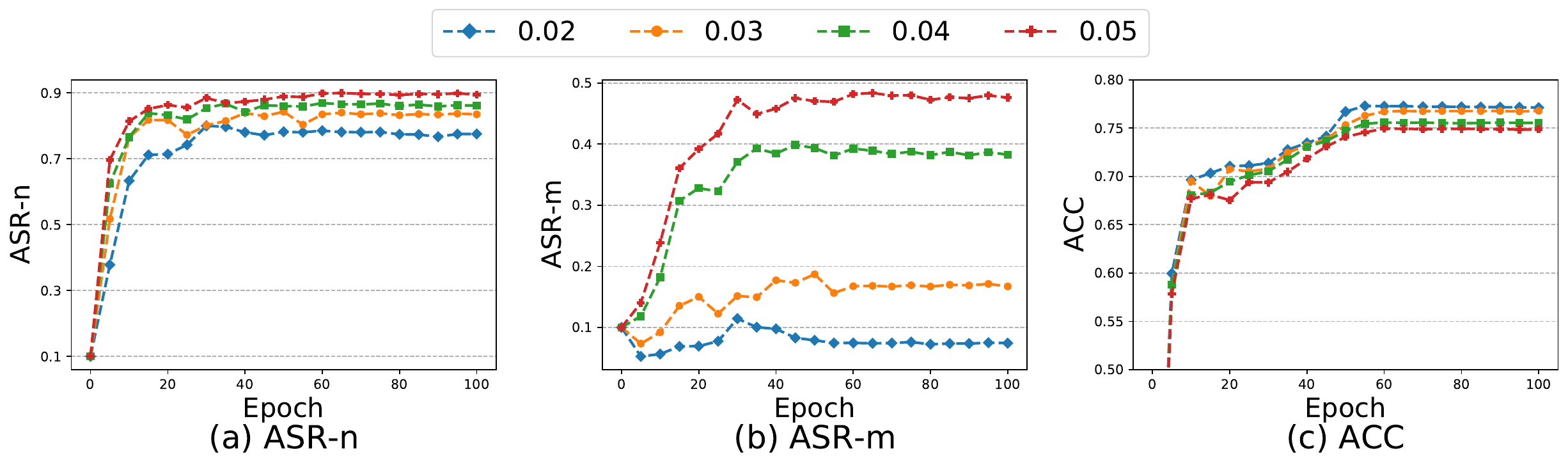} 
	\vspace{-10px}
	\caption{Performance with different poisoning ratios}
	\label{fig:5}
	\vspace{-10px}
\end{figure*}

\textbf{Evaluation Metrics.}
Let $\mathcal{D}^{test}$ denote the test set, and $\hat{\mathcal{D}}^{test}$ denote the set of samples in $\mathcal{D}^{test}$ whose ground-truth label is not equal to the backdoor class $y^T$.
Subsequently we can divide the images of the samples in $\hat{\mathcal{D}}^{test}$ into two categories: $\mathcal{F}'_1=\{\mathbf{X}|(\mathbf{X},y)\in\hat{\mathcal{D}}^{test}\land f(\mathbf{X})=1\}$ and $\mathcal{F}'_0=\{\mathbf{X}|(\mathbf{X},y)\in\hat{\mathcal{D}}^{test}\land f(\mathbf{X})=0\}$.
We call the images in $\mathcal{F}'_1$ as the natural poisoned images because they naturally contain the backdoor trigger.
Let $\Phi$ denote the target model.
In general we observe our attack performance in terms of three metrics as following:
(i) accuracy (\textbf{ACC}), the classification accuracy of the target model on $\mathcal{D}^{test}$ (\textit{i.e.,} $\frac{|\{(\mathbf{X},y)\in\mathcal{D}^{test}|\Phi(\mathbf{X})=y\}|}{|\mathcal{D}^{test}|}$);
(ii) natural attack success rate (\textbf{ASR-n}), the proportion of the target model outputs equal to $y^T$ when the images in $\mathcal{F}'_1$ are the inputs (\textit{i.e.,} $\frac{|\{\mathbf{X}\in\mathcal{F}'_1 | \Phi(\mathbf{X})=y^T\}|}{|\mathcal{F}'_1|}$);
(iii) Manual Attack Success Rate (\textbf{ASR-m)}, the proportion of the target model outputs equal to $y^T$ when the generated poisoned images based on the images in $\mathcal{F}'_0$ are the inputs (\textit{i.e.,} $\frac{|\{\mathbf{X}\in\mathcal{F}'_0 | \Phi(\mathbf{t}(\mathbf{X}))=y^T\}|}{|\mathcal{F}'_0|}$).
Note that to prevent the influence of randomness all our evaluation metrics are averaged over multiple replicated experiments.

\textbf{Parameter Settings.}
Unless otherwise specified, the hyper-parameters in our attacks are set as following: $\beta=0.05$, $\lambda=1.0$, $S=3$ and $M=128$.
And the backdoor class $y^T$ is set to $6$ on both datasets.
In our attacks with the learning strategy, we utilize the Adam optimizer with $learning\_rate=0.01$.
The kernel matrix $K^{S\times S}$ is trained for $1$ epoch on MNIST and $3$ epochs on CIFAR10, respectively.
In the target model training, we utilize the SGD optimizer with $learning\_rate=0.01$, $momentum=0.9$ and $weight\_decay=0.0001$.
The number of training epochs is $5$ for FCNN and $100$ for ResNet20 and SimpleViT.
The batch size is $128$.
At the end of training all of the models reach convergence.
Note that since the bottom-right $3\times 3$ squares of most images in MNIST are completely blank, we set a larger kernel size $S=16$ for MNIST.
We conduct our experiments on a Ubuntu Server with 8 NVIDIA RTX A5000 GPUs, 64-bit 16-core Intel(R) Xeon(R) Gold 5218 CPU @ 2.30GHz and
256 GBs of RAM.

\subsection{Attack Effectiveness (RQ1)}
To verify the generality of our attack effectiveness, we perform three groups of our attacks with the randomizing strategy on MNIST and CIFAR10.
Specifically, on MNIST we adopt FCNN as the target model.
On CIFAR10 we adopt ResNet20 and SimpleViT as the target model respectively.

Figure~\ref{fig:1} shows the evaluation metrics in each epoch under three experimental settings.
On MNIST x FCNN, the target model attains its best ACC in the third training epoch, and at this point ASR-n and ASR-m achieve $0.611$ and $0.766$ respectively;
On CIFAR10 x ResNet20, the best ACC is achieved in the $60$-th training epoch, and at this point ASR-n and ASR-m are $0.899$ and $0.482$ respectively;
On CIFAR10 x SimpleViT, the best ACC is achieved in the $50$-th training epoch, and at this point ASR-n and ASR-m are $0.627$ and $0.391$ respectively.
The results under all three experimental settings show that our attacks significantly increase both two attack success rates (ASR-n and ASR-m) compared to their initial value $0.1$.
It means if the target model is applied to real life, the attacker has a high probability of being able to activate the backdoor in it through the natural poisoned images or the generated poisoned images.

From the above observations, we can conclude our attacks are effective across various datasets and target models.

\subsection{Impact of Poisoning Ratio (RQ2)}
To explore the impact of the poisoning ratio $\beta$ on our attack performance, we experiment on CIFAR10 with ResNet20 as the target model.
We carry out our attacks with the randomizing strategy, and $\beta$ is set to $0.02$, $0.03$, $0.04$ and $0.05$ respectively.
From the experimental results shown in Figure~\ref{fig:5}, we can summarize the following three points:
(i) the higher $\beta$ causes the higher ASR-n and ASR-m, meaning the more severe damage of our attacks;
(ii) the higher $\beta$ results in the lower ACC, although the drop in ACC is quite slight;
(iii) when $\beta$ is extremely small ($\beta=0.02$), most of our generated poisoned images are ineffective, but the natural poisoned images still have a high probability of successfully activating the backdoor.

\begin{figure}[t]
	\centering
	\hspace{-2em}
	\includegraphics[width=0.6\linewidth]{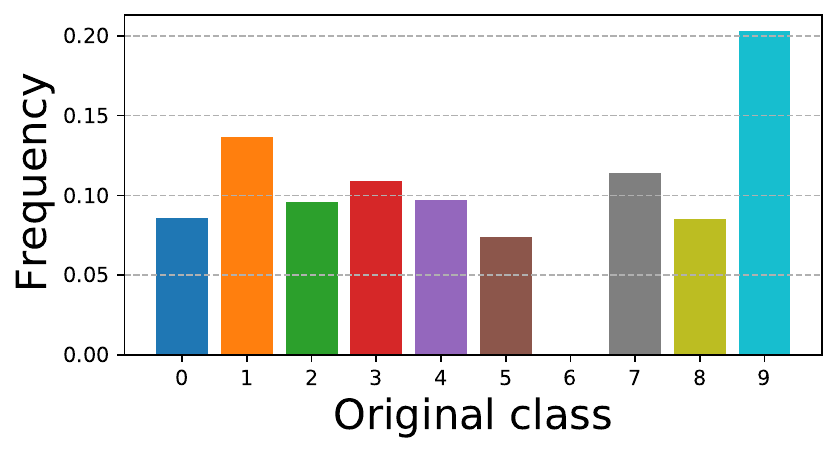} 
	\vspace{-10px}
	\caption{Original class distribution of falsified-label images}
	\vspace{-15px}
	\label{fig:2}
\end{figure}

\subsection{Attack Stealthiness (RQ3)}
We conduct our attacks with the strategy of randomizing on CIFAR10, and investigate the stealthiness of our attacks from two aspects:
(i) in the label poisoning phase, whether there are any patterns that could lead to the falsified labels being noticed;
(ii) in the trigger adding phase, whether there is perceptible difference in the images before and after adding the backdoor trigger.

Regarding the first aspect we consider that our attacks will be more detectable if most of the label-falsified images are of the same class.
Hence, we explore the distribution of the original clean class of the label-falsified images, as shown in Figure~\ref{fig:2}.
Note that due to the fact our backdoor class is set to $6$, we do not need to falsify the labels of the images whose original clean label is $6$.
From the figure we can draw the conclusion that, although the frequency of the original class $9$ is relatively a little high, the overall distribution is quite random.
And because of the randomness, it is extremely difficult for the defenders to notice the poisoned samples which are only a small fraction of the whole training set.

As for the second aspect, we directly present the original clean images and the corresponding generated poisoned images in Figure~\ref{fig:3}.
The original clean images are shown at the top, and the corresponding generated poisoned images are shown at the bottom.
Note that these original clean images are randomly selected from the test set, and all their labels are not equal to the backdoor class.
From the figure we find it difficult to distinguish the original clean images and the generated poisoned images.

In summary, our attacks are of high stealthiness.

\subsection{Comparison between Two Strategies (RQ4)}
To investigate the role of the learning strategy, we respectively perform our attacks with the strategies of randomizing and learning on CIFAR10 with ResNet20 as target model.

\begin{figure}[t]
	\centering
	\includegraphics[width=1\linewidth]{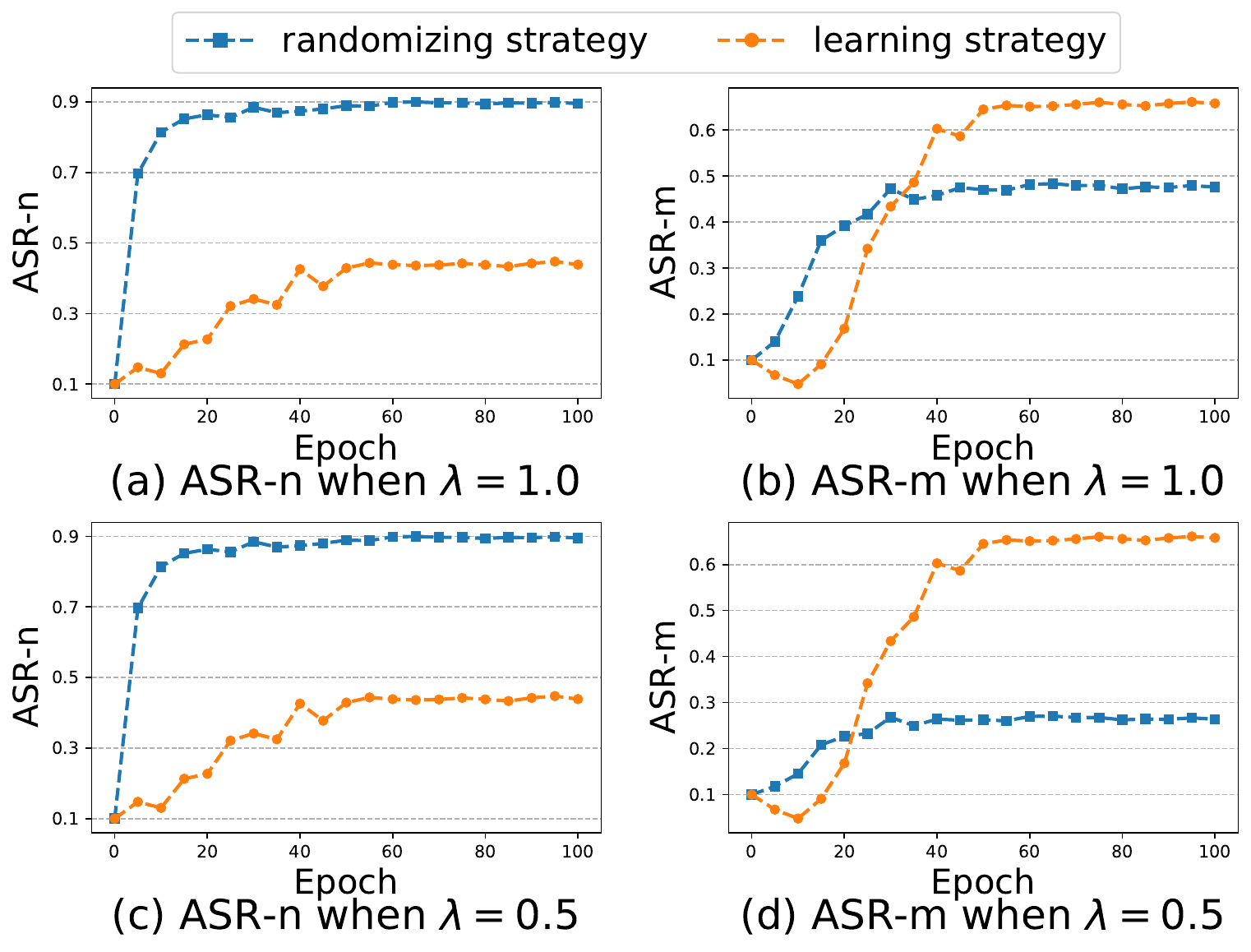} 
	\vspace{-20px}
	\caption{ASR under different attack strategies}
	\label{fig:4}
 \vspace{-15px}
\end{figure}

In Figure~\ref{fig:4}, (a) and (b) respectively show ASR-n and ASR-m under our two attack strategies.
As the figure shown, compared to the strategy of randomizing, the strategy of learning has lower ASR-n, but can achieve higher ASR-m.
We argue the reasons for this phenomenon are as following:
(i) the pattern of the function $f$ based on the learned kernel $K^{S\times S}$ is more complicated and is more difficult for the target model to precisely capture, hence the lower ASR-n;
(ii) the learned kernel $K^{S\times S}$ leads to better feature sensitivity, which means that the backdoor can be more fully activated in the trigger adding phase, hence the higher ASR-m.

When we more strictly limit the $\ell_2$-norm of the difference between clean images and poisoned images, the above phenomenon becomes more evident.
We keep the other settings unchanged but set a smaller value of the $\ell_2$-norm limit $\lambda=0.5$. 
The experimental results are shown in Figure~\ref{fig:4} (c) and (d).
From these results, we can draw the following conclusions:
(i) the $\ell_2$-norm limit $\lambda$ has no influence on ASR-n;
(ii) when $\lambda$ is smaller, the improvement of the learning strategy for ASR-m is greater.

%% file: section/conclusion.tex
\section{Conclusion and Future Work}
In this paper, we propose clean-image backdoor attacks, in which the attacker only poisons the training labels.
More specifically, we design a kernel based binary classification function to divide the training images into two groups.
According to the kernel based function, the attacker falsifies a limited proportion of the training labels in the label poisoning phase, and generates the poisoned images based on the clean images to activate the backdoor in the trigger adding phase.
The experimental results show the effectiveness, stealthiness and generality of our attacks, which indicates that our clean-image attacks are quite practical.

This paper is the first to conduct clean-image backdoor attacks and explore the attacks from various angles, which is of great significance.
However, there is still some room for improvement.
When some complicated data augmentation methods~\cite{shorten2019survey,zhong2020random,ratner2017learning,chen2020gridmask} are adopted in the training, the shortcut pattern designed by attacker may be severely disrupted, hence the poor attack effectiveness.
Besides, the resistance of our attacks to existing backdoor detection methods~\cite{gao2019strip,selvaraju2017grad,chen2018detecting,xu2023universal} and defense methods~\cite{liu2018fine,qiu2021deepsweep,wang2019neural} deserves more discussion.
Therefore, in our future work we are going to focus on improving our attack robustness to data augmentation and exploring the methods to resist our attacks.